%% file: main.tex
\documentclass{article}

\PassOptionsToPackage{numbers, compress}{natbib}

\usepackage[utf8]{inputenc} 
\usepackage[T1]{fontenc}    
\usepackage{hyperref}       
\usepackage{url}            
\usepackage{booktabs}       
\usepackage{amsfonts}       
\usepackage{nicefrac}       
\usepackage{microtype}      
\usepackage{xcolor}         

\usepackage{amsmath}
\usepackage{amssymb}
\usepackage{mathtools}
\usepackage{amsthm}
\usepackage{algorithm}
\usepackage{algorithmic}
\usepackage{tikz}
\usepackage{enumitem,kantlipsum}

    \usepackage[final]{neurips_2024}

\usepackage[utf8]{inputenc} 
\usepackage[T1]{fontenc}    
\usepackage{hyperref}       
\usepackage{url}            
\usepackage{booktabs}       
\usepackage{amsfonts}       
\usepackage{nicefrac}       
\usepackage{microtype}      
\usepackage{xcolor}         
\usepackage{subfig}
\usepackage{graphicx}
\usepackage{caption}
\usepackage{subcaption}

\input{macros}

\title{Brain in the Dark: Design Principles for Neuromimetic Inference under the Free Energy Principle}

\author{
  Mehran Hossein Zadeh Bazargani    \footnotemark[1]\hspace{0.4em}\\
  Insight SFI Research Centre, University College Dublin \& \\ School of Psychological Sciences, Monash University \\
  \texttt{mehran.hosseinzadehbazargani@monash.edu} \\
  \vspace{-4ex} 
  \And
  Szymon Urbas\footnotemark[1] \\
  Department of Mathematics \& Statistics, \\ Hamilton Institute, Maynooth University \\
  \texttt{szymon.urbas@mu.ie } \\
  \vspace{-4ex} 
  \AND
  Karl Friston \\
  Institute of Neurology, University College London \\
  \texttt{k.friston@ucl.ac.uk } \\
}

\makeatletter
\g@addto@macro\@maketitle{\vspace{-3ex}} 
\makeatother

\begin{document}
\maketitle

\renewcommand{\thefootnote}{\fnsymbol{footnote}}

\footnotetext[1]{Equal contribution.}

\begin{abstract}
Deep learning has revolutionised artificial intelligence (AI) by enabling automatic feature extraction and function approximation from raw data. However, it faces challenges such as a lack of out-of-distribution generalisation, catastrophic forgetting and poor interpretability. In contrast, biological neural networks, such as those in the human brain, do not suffer from these issues, inspiring AI researchers to explore neuromimetic deep learning, which aims to replicate brain mechanisms within AI models. A foundational theory for this approach is the Free Energy Principle (FEP), which despite its potential, is often considered too complex to understand and implement in AI as it requires an interdisciplinary understanding across a variety of fields. This paper seeks to demystify the FEP and provide a comprehensive framework for designing neuromimetic models with human-like perception capabilities. We present a roadmap for implementing these models and a Pytorch code repository for applying FEP in a predictive coding network.
\end{abstract}

\section{Introduction} \label{introduction}
The human brain, despite being confined within the darkness of the skull, possesses a remarkable ability to interpret the world around it, understand and analyse the world \textit{out there}, plan for unseen futures, and make decisions that can alter the course of events. This extraordinary capability of the brain is conjectured to come from its function as a predictive machine, constantly inferring the hidden causes behind sensory inputs to maintain a coherent understanding of its environment. This view, which dates back to Helmholtz's idea of ``perception as unconscious inference'' \cite{helmholtz1866concerning}, and later evolved as the “Bayesian brain” hypothesis \cite{doya2007bayesian}, suggests that the brain operates as a sophisticated statistical organ. It updates its beliefs about the external world based on incoming sensory data while optimising this process through a Generative Model (GM). This GM enables the brain to infer both the dynamic states of the external environment that generate its sensory inputs, as well as the mechanisms by which these inputs are produced. Essentially, the brain continually refines its probabilistic beliefs about the hidden states of the world \cite{parr2022active}, guided by the principles of Bayesian inference \cite{bayes_theorem}.

More technically, given a sensory observation $y$, the goal of perception is to infer the most likely hidden state of the world $x$, that \textit{caused} this observation. This is achieved through the Bayes’ theorem. One of the most promising frameworks for developing brain-inspired computation is through the Free Energy Principle (FEP) \cite{friston2010free}, an information-theoretic principle which posits that the brain operates to minimise a quantity known as the Variational Free Energy (VFE). VFE provides an upper bound on the negative logarithm of the Bayesian model evidence, defined as $-\ln(p(y|M))$, where $M$ is the GM. Under certain assumptions, VFE can be defined as the difference between the sensory data that the brain predicts and what it actually receives. The principle suggests that the brain seeks to reduce this discrepancy to sustain a state of equilibrium, allowing the ``self'' to survive and persist over time in an unpredictable environment. 

Despite the foundational insights offered by FEP, applying it to neuromimetic AI is challenging because it requires an interdisciplinary understanding across fields such as dynamical system modelling (through State Space Models (SSMs)), stochastic processes, probability theory, variational calculus, and neuroscience. Thus, due to the required polymathism for pursuing this line of inquiry, only a few AI researchers work with FEP. Further limiting the wide-spread use in the AI community, the initial implementation of FEP was in Matlab\footnote{\href{https://www.fil.ion.ucl.ac.uk/spm/}{https://www.fil.ion.ucl.ac.uk/spm/}}, which is less commonly used in the AI community compared to Python or Pytorch. To address these barriers, this paper contributes the following:
\begin{enumerate}
    \item A roadmap for accurately and efficiently designing neuromimetic AI using FEP.
    \item A light and CPU-based Pytorch code repository, implementing FEP in a predictive coding (PC) network \footnote{\href{https://github.com/MLDawn/PC-network-NeurIPs-2024}{https://github.com/MLDawn/PC-network-NeurIPs-2024}}.
\end{enumerate}

The rest of the paper is as follows: Section~\ref{Inference, learning, and uncertainty estimation}, introduces VFE and model inversion; Section~\ref{various problem formulations and their implications}, provides details on various problem formulations and their implications in designing FEP-based neuromimetic AI; Section~\ref{preditive coding}, introduces PC and provides its mathematical formulation; Section~\ref{experiment} details the experiment and results. Finally, Section~\ref{conclusion} concludes the paper.

\section{Inference, learning, and uncertainty estimation}\label{Inference, learning, and uncertainty estimation}
For a neuromimetic AI model to function effectively in a dynamic and ever-changing world, it must continuously adapt to new sensory inputs. It requires a Generative Model (GM) that encapsulates its understanding of the hidden \textit{Generative Process (GP)} underlying the sensory data. The GP is not directly accessible to the model, much like how the true external world behind the skull is hidden from our brain. Thus, determining the hidden states of the world becomes an inference problem, where the model seeks to reverse-engineer the GP from observed sensory inputs. This involves \textit{model inversion}, which allows us to infer the most likely hidden states that could have generated the given sensory data. Interestingly, in the AI and machine learning community, the primary focus has often been on parameter estimation rather than hidden state estimation. 

Let $z^t$ denote the set of all quantities to be inferred at time $t$; e.g.\  hidden states and/or the GM parameters. By Bayes' theorem, the posterior distribution of $z^t$ given the observed data $y^t$, is expressed as: $p(z^t|y^t)=p(y^t|z^t)p(z^t)/p(y^t)$. The calculation of the \emph{Bayesian model evidence} term, $p(y^t)$, often involves a complex, high-dimensional integral that is generally intractable. To circumvent this difficulty, approximate Variational Inference (VI) is considered, which equates to a simpler optimisation problem. The aim is to find a surrogate distribution $q(z^t)$ that approximates the true posterior $p(z^t|y^t)$ by minimising the \emph{Variational Free Energy (VFE)} \cite{friston2010free}: 
\begin{equation}\label{VFE}
        F(q;y^t) = \underbrace{D_{\mathrm{KL}}[q(z^t)||p(z^t)]}_{\text{Complexity}}-\underbrace{\mathbb{E}_{q(z^t)}[\ln p(y^t|z^t)]}_{\text{Accuracy}}
                        = -\mathbb{E}_{q(z^t)}\left[\ln\frac{ p(y^t,z^t)}{q(z^t)}\right],
\end{equation}
where $D_{\mathrm{KL}}$ is the \textit{Kullback-Leibler} divergence. Minimising VFE serves two purposes: it approximates the model evidence, and it provides a robust criterion for selecting among different GM models. Crucially, since VFE is a functional of $q$ (i.e.~it takes in a function and returns a scalar), calculus of variation is used for the minimisation \citep[e.g.~Chapter 10 of][]{BiNa2006}. The VFE balances two opposing quantities: \textit{accuracy}, which ensures that the model's predictions closely match the observed data, and \textit{complexity}, which penalises overly complex models that might overfit. Specifically, complexity measures the extent to which the prior belief of the model about the state of the world, $p(z^t)$, will shift towards the approximate posterior belief, $q(z^t)$ after having observed, $y^t$. By minimising VFE, the model achieves an optimal trade-off between fitting the data and maintaining simplicity, adhering to the principle of Occam's razor.

The inference process through VFE minimisation endows neuromimetic AI with three crucial capabilities: (i) \textbf{Parameter Estimation}: learning the parameters of the generative model that best explain the data; (ii) \textbf{Precision Estimation}: estimating the precision (inverse uncertainty)--- discussed in Section~\ref{preditive coding}---over hidden states and observations, and; (iii) \textbf{State Estimation}: inferring the hidden states that caused the observed data. All three capabilities are essential for constructing truly neuromimetic AI systems that can adapt and generalise across different contexts, much like biological neural networks. For illustration purposes, however, we focus on the scenario (iii); that is $z^t=x^t$, keeping the remaining aspects of the generative model fixed (i.e.\ fixed parameters and state/observation precision terms).

\section{Various problem formulations and their implications}\label{various problem formulations and their implications}
When designing GMs, and the method for their inversion, various problem spaces need to be considered. This section explores the different formulations and their implications for developing neuromimetic AI based on the FEP. In what follows, we discuss discrete-time, discrete-space Markov chains, and continuous-time, continuous-space random processes; other formulations are outside the scope of this paper.

\textbf{Discrete State Space Models}: In discrete state space models, the hidden state, $x^t$, can take $K_x$ possible values, whilst the observed datum, $y^t$, can take $K_y$ values; $t=0,1,\ldots$. As the hidden states and observations are categorical variables, the likelihood also follows a categorical distribution, parameterised by the $K_x\times K_y$ matrix $\mathbf{A}$: $y^t|x^t=j \sim \mathsf{Cat}(\mathbf{A}^{(j)})$, where $\mathbf{A}^{(j)}$ is the $j^\mathrm{th}$ row of $\mathbf{A}$ and the matrix entries are $A_{jk} = \mathbb{P}(y^t = k|x^t = j)$; $j=1,...,K_x$,  $k=1,...,K_y$. Similarly, the transition probability for the latent states is parameterised by the $K_x \times K_x$ matrix $\mathbf{B}$: $x^{t + 1} | x^t=j \sim \mathsf{Cat}(\mathbf{B}^{(j)})$, with entries $B_{jk} = \mathbb{P}(x^{t+1} = k|x^t = j)$. The initial prior probabilities of these states are encoded in a $K_x$-dimensional vector $D$ expressed as a categorical distribution, that is, $x^0\sim\mathsf{Cat}(D)$. The full prior over a sequence of hidden states up to some time $\tau$, denoted as $x^{0:\tau}$ is $p(x^{1:\tau})=p(x^1)\prod_{t=1}^{\tau-1} p(x^{t+1}| x^t)$. This formulation forms the basis of the Hidden Markov Model (HMM), commonly used for inference tasks \citep[e.g.][]{Rabi1989}, but it may also be used for learning which is outside the scope of this paper; see Fig.~\ref{discrete_hmm} in Appendix~\ref{app:HMM} for a visual representation.
While this formulation assumes a finite state space, it can be extended to infinite state spaces with appropriate assumptions about the generative process \citep[e.g.][]{Brun2013}.

\textbf{Continuous State Space Models}: In continuous state space models, both hidden states and observations take continuous real values. The temporal evolution of the hidden states $x^t$, and their relationship to observations $y^t$, are defined through a system of stochastic differential equations parameterised by $\theta$. To reduce notational burden, $\theta$ represents the collection of separate parameters for each part of the model; the parameters are sometimes referred to as \textit{causes}. The equations are
\begin{equation}\label{eq:ssm}
\dot{x}^t = f(x^t, \theta) + \omega_x^t~~~\mbox{and}~~~    y^t = g(x^t, \theta) + \omega_y^t,
\end{equation}
evolutionwhere  $\dot{x}^t$ is the first-order time derivative of the hidden state $x^t$, representing the rate of change (i.e.\ velocity) of the hidden state. Here, $\omega_x^t$ and $\omega_y^t$ represent the random fluctuations corresponding to the states and observations, respectively; the two random processes are assumed to be independent (e.g.~\citep{friston2010generalised}). In the most basic case, these could be Wiener processes \citep[e.g.][]{CoMi1965}, with independent Gaussian increments, but other smoother processes such as the Mat\'ern process could be used here \citep[e.g.][]{HaSa2010}. The first equation describes the evolution of hidden states over time through a deterministic function $f(x^t, \theta)$ and stochastic fluctuations  $\omega_x^t$, where the evolution of the hidden states can be modelled as differential equations, e.g.~the change from some $t_0>0$ to some other $t_1>t_0$ comprises infinitesimally small increments in time. The second equation expresses how the observations are believed to be generated from the hidden state through a deterministic function $g(x^t, \theta)$ and stochastic fluctuations $\omega_y^t$. Interestingly, if we assume the fluctuations to be zero-mean Gaussian processes, these two equations form a GM that underwrites the \textit{Kalman-Bucy filter} \citep[e.g.][]{ruymgaart2013mathematics} in the engineering literature. Crucially, even though we may be observing this continuous state space model at discrete times, the underlying dynamics of the system are continuous in time (e.g.~the evolution of the hidden states, VFE minimisation, etc.). Here, the hidden states and their velocities are collapsed into one latent variable $\tilde{x}^t=\{\dot{x}^t, x^t\}$---of interest is the approximate posterior $q(\tilde{x}^t)$. The standard VFE can be derived and minimised, during the time intervals between observations. More specifically, after observing $y^t$, we can minimise the integration of point estimates of VFE along a continuous time interval until the next observation $y^{t+\Delta}$, $\Delta>0$. This quantity is called \emph{Free Action} and is defined as $\overline{\mathcal{A}}[q(\tilde{x})]=\int_{t}^{t+\Delta} F(q(\tilde{x}^s);y^s) \,\mathrm{d}s$, and it is an upper bound on the accumulated surprise, $
-\int_{t}^{t+\Delta} \ln(P(y^s)) \,\mathrm{d}s$, over the same time period. In practice, one does not observe the data on the continuum and as such $y^t$ is used to approximate $\{y^s, s\in[t,t+\Delta)\}$ in the integrand. By minimising $\overline{\mathcal{A}}$ in-between observations, the generative model is constantly minimising VFE along a path of length $\Delta$, and thus continuously striving to improve the approximate estimation of the posterior over the hidden states (and potentially parameters, which is outside the scope of this paper). Interestingly, it is possible and indeed biologically plausible to relax the assumption of independent increments of $\omega_x^t$ and $\omega_y^t$, which can endow the GM with a more agile tracking ability of the external world (See. Appendix~\ref{app:A note on the Generalised Coordinates of Motion})

\section{Predictive coding}\label{preditive coding}
To maintain stability (i.e.\ homeostasis), and ensure survival, biological systems like the brain must continuously minimise fluctuations or entropy in their internal and external states. This process is akin to minimising the brain's ``surprise'' about its sensory states, which, from a statistical perspective, translates to maximising the Bayesian model evidence for its sensory input---a process known as Bayesian filtering. Predictive coding \cite{friston2005theory,auksztulewicz2016repetition} is a prominent and neurobiologically plausible approach to Bayesian filtering, which frames the brain's function as a constant interplay between prediction and error correction.
Under the predictive coding framework, the brain is seen as a hierarchical generative model that optimises its internal model of the world by minimising prediction errors. These errors are the differences between the brain's predictions (top-down signals) and the actual sensory inputs (bottom-up signals). The brain accomplishes this through a two-fold process: first, by generating top-down predictions about sensory inputs, and second, by calculating the prediction errors (bottom-up signals) that serve to update these predictions.
The VFE provides a mathematical approximation for the Bayesian model evidence, which, under certain conditions, is equivalent to precision-weighted prediction errors. This is achieved using the Laplace approximation, a method that approximates complex model distributions with simpler Gaussian distributions. Inferring under the variational paradigm, one arrives at Variational Laplace (VL) \cite{ZEIDMAN2023120310}, allowing for efficient computation and optimisation of VFE in a biologically plausible manner.
In this framework, perception is conceptualised as the minimisation of prediction errors through the continual updating of expectations that propagate down the cortical hierarchy. Predictions flow downward from deeper cortical layers to more superficial ones, while the resulting prediction errors travel upward, refining the brain's expectations and improving future predictions. In essence, the brain functions as a self-correcting system, constantly seeking to reduce the discrepancies between its expectations and sensory reality, thereby optimising its internal model of the world.
Mathematically, predictive coding can be modelled as a hierarchical state space model, where each of the $L$ layers of the hierarchy represents a level of abstraction:
\begin{equation}\label{eq:ssm_hierarchical}
    \begin{array}{c}
        \begin{minipage}{0.25\textwidth}
            \begin{equation*}
            \begin{aligned}
                \dot{x}_1^t &= f_1\left(x_1^t, \theta_1^t\right) + \omega_{x,1}^t \\
                y^t &= g_1\left(x_1^t, \theta_1^t\right) + \omega_{\theta,1}^t
            \end{aligned}
            \end{equation*}
        \end{minipage}
        ,~~~~\mbox{and}~~~~
        \begin{minipage}{0.28\textwidth}
            \begin{equation*}
            \begin{aligned}
                \dot{x}_l^t &= f_l\left(x_l^t, \theta_l^t\right) + \omega_{x,l}^t \\
                \theta_{l-1}^t &= g_l\left(x_l^t, \theta_l^t\right) + \omega_{\theta,l}^t
            \end{aligned}
            \end{equation*}
        \end{minipage},
        \begin{minipage}{0.2\textwidth}
            \begin{equation*}
            \begin{aligned} 
                l=2,...,L;
            \end{aligned}
            \end{equation*}
        \end{minipage}
    \end{array}
\end{equation}
where $\omega_{x,l}^t$ and $\omega_{\theta,l}^t$, are the random fluctuations in $\mathit{layer}_l$, which if, we assume to be zero-mean Gaussian processes, lead to approximately Gaussian conditional distributions: $\dot{x}_l^t | x_l^t, \theta_l^t\sim \mathsf{N}(f_l(x_l^t, \theta_l^t), \Pi_{x_l^t}^{-1})$ and $\theta_{l-1}^t | x_l^t, \theta_l^t\sim \mathsf{N}(g_l(x_l^t, \theta_l^t), \Pi_{\theta_l^t}^{-1})$, where precision terms $\Pi_{x_l^t}$ and $\Pi_{\theta_l^t}$ are based on the assumed variance structure of the random fluctuations. Indeed, $\mathit{layer}_l$ infers the most likely distribution of the hidden state, $\tilde{x}_{l-1}^t=	\{\dot{x}_{l-1}^t,x_{l-1}^t\}$, in $\mathit{layer}_{l-1}$ by minimising VFE (in a purely local and Hebbian sense between layers $\mathit{layer}_{l-1}$ and $\mathit{layer}_l$). The same VFE minimisation approach takes place between each pair of consecutive layers in a local fashion. 
At the bottom of the hierarchy, the first layer resembles the sensory epithelia, tasked with inferring the hidden states of the external world based on noisy sensory signals. As each layer minimises its VFE independently, the entire generative model is effectively inverted, achieving hierarchical inference of the hidden states that cause sensory observations—this is essentially the process of perception.
The communication between layers relies on the parameters $\theta$, which enable consecutive layers to predict the states in adjacent layers. This hierarchical message-passing scheme reflects the brain's ability to integrate and process information across different levels of abstraction. Simplifying to a single-layer PC network, the model is mathematically equivalent to a basic state space model as described in Section~\ref{various problem formulations and their implications}. For a detailed explanation of neuronal message passing within a one-layer PC network, see Appendix ~\ref{app:PC message passing}.


\section{Experiments with a one-layer PC model \& results}\label{experiment}
We present experimental results demonstrating how a simple one-layer PC network can infer the hidden states of the external world from noisy sensory inputs; implementation details are in the provided CPU-based Pytorch code repository. All experiments are performed on a personal laptop---with \textit{Intel(R) Core-i9 CPU} and \textit{16GB (RAM)}. The pseudo-code is provided in the Appendix \ref{app:pseudocode}.

\textbf{The GP}: We consider a GP (i.e.\ the ``true'' external world) which is modelled using Lotka-Volterra process \cite{wangersky1978lotka}, describing dynamics of a biological system in which two species interact, typically predator and prey: $\mathrm{d} x^t[0]/\mathrm{d}t = \alpha x^t[0] - \beta x^t[0]x^t[1]$ and $\mathrm{d}x^t[1]/\mathrm{d}t = -\gamma x^t[1] + \delta x^t[0]x^t[1]$, where $x^t[0]$ and $x^t[1]$ are the populations of the prey and predator at time $t$, respectively; $\alpha$ defines the natural growth rate of the prey when no predators are present and $\beta$ is the rate at which predators kill the prey. The coefficients were set to $(\alpha,\beta,\gamma,\delta)=(0.7, 0.5, 0.3, 0.2)$. Euler's method was used to numerically solve this system of ODEs over a total time of $T=100$, with the initial conditions $(x^0[0], x^0[1]) = (1.0, 0.5)$ and a time-discretisation of $\varepsilon=0.1$. The solution paths $x^t[0]$ and $x^t[1]$ are each of length 1000 ($=T/\varepsilon$) and are presented in the left panel of Fig.~\ref{fig:x and y} in Appendix~\ref{app:Lotka-Volterra GP and observations}. The observations, $y^t[0]$ and $y^t[1]$---in the right panel of Fig.~\ref{fig:x and y} in Appendix~\ref{app:Lotka-Volterra GP and observations}---are generated by adding a coloured (i.e.\ correlated) noise to the solutions, $x^t[0]$ and $x^t[1]$, of the generative process, independently; the noise is a Wiener process convolved with a smoothing kernel (details in the code).

\textbf{The GM}: The generative model follows the form presented in Eq.~\eqref{eq:ssm}, with two different kinds of flows, $f$, considered. For the first model $M_1$, the flow is a linear pullback attractor: $f_1(x)= - A (x - \varphi)$, with $A=\begin{bsmallmatrix}
         0.5 &0\\
         0 & 0.5
\end{bsmallmatrix}$ and $\varphi= (1, 1)^\top$. For the second model $M_2$, the flow follows non-linear trigonometric dynamics: $f_2(x)=(\sin{x[0]},\sin{x[1]})^\top$. We choose the observation model for both $M_1$ and $M_2$ to be the identity mapping, $g_1(x)=g_2(x)= x$. Assuming the random fluctuations,  $\omega^t_x$ and $\omega^t_y$ from Eq.~\eqref{eq:ssm}, to be zero-mean Gaussian processes, the GM likelihood is constructed from the conditional probability density functions of model variables: $p(\dot{x}^t | x^t, \theta^t)=p_N(\dot{x}^t;f(x^t, \theta^t), \Pi_{x^t}^{-1})$ and $p(y^t | x^t, \theta^t)=p_N(y^t ; g(x^t, \theta^t), \Pi_{y^t}^{-1})$; where $p_N(\,\cdot\,;\mu,\Sigma)$ is the density function of a (Gaussian) $\mathsf{N}(\mu,\Sigma)$ variable. 
To aid in the inference of the latent velocity $\dot{x}$, it is typical to use a regularising prior $\nabla f(x)\dot{x}\sim\mathsf{N}(0,\Pi_x)$ which draws $\dot{x}$ towards zero to prevent potential over-fitting \citep[e.g.][]{Fris2008,HeMi2024}. The likelihood is then obtained from the approximate distribution of the error terms: $\varepsilon_x(\tilde x^t) =\left(
    \dot{x}^t - f({x}^t),\nabla f(x)\dot{x}
\right)^\top$
and $\varepsilon_y(\tilde x^t) = y^t - g(x^t)$, where $\tilde x^t = ( x^t, \dot{x}^t)^\top$. The goal of the inference is to identify the posterior distribution over $\tilde x^t$. Here, the precision terms are fixed $\Pi_{x^t} = \Pi_{y^t} = \begin{bsmallmatrix}
         1 &0\\
         0 & 1
\end{bsmallmatrix}$; however, these can be estimated by minimising the  VFE \citep[e.g.][]{Fris2008}.    For a detailed treatment of the functional form of VFE and its use in the inference for our experiments, see Appendix~\ref{app:VFE functional and mfa}.

\begin{table}[t]
    \centering
    \begin{tabular}{c c c c}
      \hline
      \multicolumn{1}{c}{GM} & \multicolumn{1}{c}{State flow $f(x,\theta)$} & \multicolumn{1}{c}{Free Action} & \multicolumn{1}{c}{MSE Loss} \\
      \hline
      $M_1$ & pullback & 576.98 & 0.53 \\
      \hline
      $M_2$ & trigonometric & \textbf{423.21} & 0.52 \\
      \hline
    \end{tabular}
    \caption{State inference experiment results for a Lotka-Volterra GP, using 2 flavours of GMs, $M_1$ and $M_2$, with pullback attractor and trigonometric state flow dynamics, respectively.}
    \label{tab:results}
\end{table}
The results of hidden state inference using two GM versions of a simple one-layer PC network, $M_1$ and $M_2$, with different state flow dynamics, $f_1$ and $f_2$ are summarised in Table.\ref{tab:results}. We can see that the non-linear nature of $f_2$ has resulted in a much lower free action in $M_2$, rendering it superior to $M_1$. For the sake of completeness, we have also presented the Mean Squared Error (MSE), which is measured by $\mathrm{MSE}(\tilde{x},\widehat{\tilde{x}})=\frac{\sum_{n=1}^{N}(\tilde{x}_n - \widehat{\tilde{x}}_n)^2}{1000}$, where $\tilde{x}$ and $\widehat{\tilde{x}}$ are the true state of the world and their posterior estimates, respectively, and $N=1000$ denotes the length of the trajectory in $\tilde{x}$. Caution must be taken in that it is the free action that matters and other measures such as MSE---that solely focus on accuracy and ignore model complexity---should not be consulted on their own, due to the risk of over-fitting, as discussed in Section~\ref{Inference, learning, and uncertainty estimation}. For a visualisation of the inferred states by $M_1$ and $M_2$ generative models, the evolution of free action during inference under each model, and how Bayesian model selection can be used to pick the best model, see Appendix~\ref{app:experiments}. Finally, the actual generative power of $M_1$ and $M_2$ models are illustrated in Appendix \ref{app:generative_power}.

\section{Conclusion}\label{conclusion}
Neuromimetic AI aims to endow traditional AI models, such as deep learning, with brain-like neuronal message-passing and human-like reasoning. The FEP is one of the most promising directions for accomplishing this. Unfortunately, due to its mathematically challenging and multi-disciplinary nature, pursuing the FEP route to neuromimeticism, understanding it and of course implementing it, remain a challenging task for researchers. This paper provides a detailed account of the design principles of neuromimetic AI models using FEP, which is applied in PC networks. Last but not least, we provide a Pytorch code repository for an exact implementation of a PC network based on FEP, which mimics human perception.
\bibliographystyle{unsrt}
\section*{Acknowledgements}
Mehran Hossein Zadeh Bazargani is supported under the European Union’s Horizon 2020 research and innovation programme under the Marie Skłodowska-Curie grant agreement No. 101034252. Szymon Urbas is grateful for the financial support of Science Foundation Ireland (SFI) and the Department of Agriculture, Food and Marine on behalf of the Government of Ireland under grant No. 16/RC/3835 - VistaMilk. Karl Friston is supported by funding from the Wellcome Trust No. 203147/Z/16/Z.

\bibliography{references}
\appendix
\onecolumn

\section{HMM for inference/learning}\label{app:HMM}
\begin{figure}[h]
\centering
\includegraphics[width=.5 \textwidth]{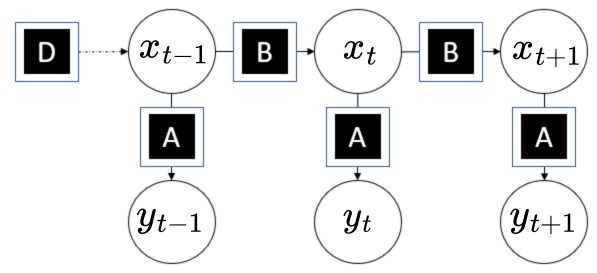}
\caption{A Hidden Markov Model (HMM) for inference  \cite{parr2022active}.}
\label{discrete_hmm}
\end{figure}

The HMM in Fig.~\ref{discrete_hmm}, represents the evolution of a sequence of hidden states, $x^t$, over time; $t$ here is on a discrete domain, e.g. $t=0,1,2,...$. At each time step, $t$, a hidden state emits an observation, $y^t$, and the state at any one time depends \textit{only} on the state at the previous time where this dependency is encoded in the matrix \textbf{B}. The initial prior probability regarding the hidden state is encoded in the vector $D$ (not to be confused with the derivative operator in Appendix~\ref{app:pseudocode}), and finally, the matrix \textbf{A} encodes the likelihood distribution of generating outcomes under each state. Here, it is assumed that the parameters of the generative model are learned and we are only interested in inferring the hidden states.

\section{A note on the generalised coordinates of motion}\label{app:A note on the Generalised Coordinates of Motion}
Importantly, random fluctuations in the data-generating mechanism in the GP, $\omega$, are generally assumed to have uncorrelated increments over time (i.e.~Wiener assumption), however, in most complex systems (e.g.~biological systems)---where the random fluctuations themselves are generated by some underlying dynamical system---these fluctuations possess a certain degree of smoothness. Indeed, by relaxing the Wiener assumption and imposing smoothness on the model functions $f$ and $g$, we have the opportunity to not only consider the rate of change of the hidden state and the observation but also their corresponding higher-order temporal derivatives (i.e.~acceleration, jerk, etc.); see, for example, \cite{friston2010generalised}. The resultant pair of $\{x^t, \dot{x}^t, \ddot{x}^t,...\}$ and $\{y^t, \dot{y}^t, \ddot{y}^t,...\}$ are called the \textit{generalised coordinates of motion} \cite{balaji2011bayesian}, which provide an opportunity for further capturing the dynamics that govern the evolution of the hidden states and observations. An estimated trajectory over time can be calculated using a Taylor series expansion around the present time, which results in a function that can extrapolate to the near future as well as the recent past. This can be of particular interest when applying the inference scheme to prediction tasks---the information contained in the generalised coordinates allows for more accurate propagation of the dynamics into the future. In the examples covered in this paper, we only consider velocity when simulating the generative model forward in time, but the inclusion of, for example, acceleration could further improve the results.

\section{The neuronal message passing in a one-layer PC network}\label{app:PC message passing}
In this appendix, we describe how a one-layer predictive coding (PC) network updates its beliefs about the state of the world and its dynamics through neuronal message passing. The model uses a combination of top-down predictions and bottom-up error signals to refine its internal beliefs about the hidden states of the world and their temporal dynamics.

\begin{enumerate}[wide, labelindent=0pt]

\item  \textbf{Top-down prediction}: The expectations of the model about the current state of the world and its dynamics are represented by $\mu_x$ (the state estimate) and $\dot{\mu}_x$ (estimated rate of change of the state), respectively. Based on these expectations, the model generates two types of predictions:
\begin{itemize}
\item The prediction for the observation, $y^t$, at time $t$ which is represented as $g(\mu_x, \theta)$.
\item The prediction for the state’s rate of change, $\dot{\mu}_x$, which is represented as $f(\mu_x, \theta)$.
\end{itemize}
These top-down predictions are based on the model’s current beliefs and its parameters, $\theta$.

\item \textbf{Bottom-up error propagation}: Next, the model compares its predictions with the actual observations. This leads to the generation of two types of prediction errors:
\begin{itemize}
\item     The error in predicting the observation, denoted as $\varepsilon_y(\mu_x) = y^t - g(\mu_x, \theta)$ and,
\item     The error in predicting the state’s rate of change, denoted as:  
\begin{align*}
\varepsilon_x(\mu_x) = (\dot{\mu}_x - f(\mu_x, \theta), -\nabla f(\mu_x, \theta)\dot{\mu}_x)^\top.
\end{align*}

\end{itemize}

These errors signal how well the model’s predictions align with the actual input, providing feedback for updating the model’s beliefs.

\item \textbf{Gradient-based belief update (GM inversion)}: The model updates its beliefs by adjusting its estimations of the hidden state, $\mu_x$, and the dynamics of the hidden state, $\dot{\mu}_x$, through the process of minimising the VFE, $F(q;y^t)$ where $q$ is an approximating distribution parameterised by $\tilde{\mu}_x = (\mu_x,\dot\mu_x)^\top$.

(i) \textit{Updating belief over the state}: If we ignore the dynamics (i.e., rate of change), the belief update for $\mu_x$ is based on the gradient of the VFE w.r.t $\mu_x$, and the update rule follows a standard gradient descent approach: 
\begin{align*}
    \mu_x^{i+1}\leftarrow\mu_x^{i} -\eta\nabla_{\mu_x} F(q;y^t)|_{\mu_x}
\end{align*}

where $\eta$ is the learning rate, and $i$ indicates the current iteration. 

However, if we incorporate the model’s expectations about the world’s dynamics $\dot{\mu}_x$, and we should, the previous belief update rule becomes: 
\begin{align*}
\mu_x^{i+1}\leftarrow\mu_x^{i} + \eta(\dot{\mu}_x -\left.\nabla_{\mu_x} F(q;y^t)\right|_{\mu_x}),    
\end{align*}
 
where the term $\eta \dot{\mu}_x$ serves as a momentum term in updating $\mu_x$. This momentum reflects the model’s belief about how fast the state is changing, leading to a smoother update.
By rearranging this update rule, we get: 
\begin{align*}
\frac{\mu_x^{i+1}-\mu_x^i}{\eta} =\dot{\mu}_x -\nabla_{\mu_x} F(q;y^t)|_{\mu_x}.    
\end{align*}
In the continuous-time limit (i.e., as $\eta \to 0$), this becomes a continuous curve, parameterised by $s>0$, which satisfies the following ordinary differential equation (ODE):
\begin{align*}
\frac{\dd}{\dd s}\mu_x^{(s)} =\dot{\mu}_x^{(s)} -\nabla_{\mu_x} F(q;y^t)|_{\mu_x^{(s)}},    
\end{align*}
where the hidden state $\mu_x$ evolves according to both the prediction errors and the expected dynamics of the world. Given an initial condition $\mu_x^{(0)}$ (i.e., the initial expectation of the GM about the hidden state of the world prior to any observation), and by integrating the ODE over time---for example using Euler's or Runge-Kutta (RK) methods---the GM updates its belief and model inversion is accomplished: $\int \frac{\dd}{\dd s}\mu_x^{(s)} \, \dd s$. 

(ii) \textit{Updating the state velocity}: Similarly, the model also updates its belief about the velocity, $\dot{\mu}_x$, of the hidden state. However, because our simple one-layer PC network does not estimate higher-order temporal derivatives (e.g., belief over the acceleration of the hidden state), the update for velocity is simplified to a standard gradient descent step: 
\begin{align*}
\dot{\mu}_x^{i+1}\leftarrow\dot{\mu}_x^{i} - \eta\left(\left.\nabla_{\dot{\mu}_x} F(q;y^t)\right|_{\dot{\mu}_x}\right).
\end{align*}

This leads to the following differential equation for the velocity update: 
\begin{align*}
 \frac{\dd}{\dd s}\dot{\mu}_x^{(s)} =-\nabla_{\dot{\mu}_x} F(q;y^t)|_{\dot{\mu}_x^{(s)}}.   
\end{align*}

In this case, the velocity is updated based purely on the prediction error without incorporating any higher-order dynamics like acceleration. Similarly, given an initial condition $\dot{\mu}_x^{(0)}$ (i.e., the initial expectation of the GM about the velocity of the world prior to any observation), and by integrating this ODE over time, the GM updates its belief: $\int \frac{\dd}{\dd s}\dot{\mu}_x^{(s)} \, \dd s$. 

In summary, the one-layer PC network updates its beliefs about both the state $\mu_x$, and the velocity $\dot{\mu}_x$, of the world through a combination of top-down predictions and bottom-up error signals. This is achieved by applying gradient descent to minimise VFE, with dynamics being incorporated as a momentum term. However, this simple model does not estimate higher-order temporal derivatives like acceleration.

Last but not least, please note that for practical purposes a \textit{Laplace-based approximation} of variational free energy $\hat F(q;y^t)$, is usually used to compute it for model inversion (See Appendix~\ref{app:VFE functional and mfa}).
\end{enumerate}

\section{State inference pseudo-code}\label{app:pseudocode}
Algorithm.~\ref{alg:pseudocode} shows the pseudo-code for the hidden state estimation problem defined in Section~\ref{experiment} where the GM is a one-layer PC network and the GP is a Lotka-Volterra process. This means that the dimensionality of hidden states $x$, and sensations $y$, is equal to 2. The pseudo-code is self-explanatory, however, in \textit{line 9}, we have a mysterious block matrix $D$ that will require further explanation.

Let $k_x$ be the number of coordinates in $\tilde{x}=(x, \dot{x})$, which is 2 (i.e., the GM estimations for position and velocity of the external world), and let $dx$ be the dimensionality of $x$, which in the case of a Lotka-Volterra process is 2. Then, we can use $D\in\Real^{k_x d_x\times k_x d_x}$, which is a \textit{block-matrix derivative operator} with identity matrices on its ﬁrst leading-diagonal, to write the belief update rule on $\mu_x$ and $\dot{\mu}_x$ in just one line, as shown in \textit{line 9} of the pseudo-code; specifically, here, $D = \begin{bsmallmatrix}
    0&1\\
    0&0
\end{bsmallmatrix}\otimes I_{d_x}$, where $\otimes$ denotes the Kronecker product and $I_{d_x}$ is a $d_x\times d_x$ identity matrix. Using $D$ is a smart way to shift the elements in $\tilde{\mu}_x$ up by $d_x=2$. This shift is crucial since, when updating its belief about the position of the world $\mu_x$, the GM will automatically use velocity $\dot{\mu}_x$ as momentum in the belief update process. Similarly, when updating its belief about the velocity of the world, $\dot{\mu}_x$, the GM will automatically use acceleration, $\ddot{\mu}_x$, as the momentum term, and so on. We will show this in action in both Eq.~\eqref{eq:update rule_1} and Eq.~\eqref{eq:update rule_2}, by expanding \textit{line 9} of the pseudo-code.

Since we know that $\tilde{\mu}_x = (\mu_x, \dot{\mu}_x)^\top$ and since $d_x=2$, then we know $\mu_x = (\mu_x[0], \mu_x[1])^\top$ and $\dot{\mu}_x = (\dot{\mu}_x[0], \dot{\mu}_x[1])^\top$ as column vectors. Then $\tilde{\mu}_x = (\mu_x[0], \mu_x[1], \dot{\mu}_x[0], \dot{\mu}_x[1])^\top$. So, \textit{line 9} can be expanded into (dropping the $t$ superscript to simplify the notation):

\begin{equation}
    \frac{\dd}{\dd t}\begin{bmatrix} \mu_x[0] \\\mu_x[1] \\\dot{\mu}_x[0] \\ \dot{\mu}_x[1] \end{bmatrix} = \underbrace{\begin{bmatrix}
    0 & 0 & 1 & 0 \\
    0 & 0 & 0 & 1 \\
    0 & 0 & 0 & 0 \\
    0 & 0 & 0 & 0
    \end{bmatrix}}_{=D} \begin{bmatrix} \mu_x[0] \\\mu_x[1] \\\dot{\mu}_x[0] \\ \dot{\mu}_x[1] \end{bmatrix}-
    \begin{bmatrix} 
    \nabla_{\mu_x[0]} \hat F(q;y_i)|_{\mu_x[0]} \\
    \nabla_{\mu_x[1]} \hat F(q;y_i)|_{\mu_x[1]} \\
    \nabla_{\dot{\mu}_x[0]} \hat F(q;y_i)|_{\dot{\mu}_x[0]} \\
    \nabla_{\dot{\mu}_x[1]} \hat F(q;y_i)|_{\dot{\mu}_x[1]} 
    \end{bmatrix}.
    \label{eq:update rule_1}
\end{equation}
We can see how the $\tilde{\mu}_x$ column vector is shifted up by $d_x=2$, after the derivative operator $D$, has been applied to it. Thus, the ODE (i.e.\ the update rule) simplifies to:

\begin{equation}
    \frac{\dd}{\dd t}\begin{bmatrix} \mu_x[0] \\\mu_x[1] \\\dot{\mu}_x[0] \\ \dot{\mu}_x[1] \end{bmatrix} = 
    \begin{bmatrix} 
    \dot{\mu}_x[0] - \nabla_{\mu_x[0]} \hat F(q;y_i)|_{\mu_x[0]} \\
    \dot{\mu}_x[1] - \nabla_{\mu_x[1]} \hat F(q;y_i)|_{\mu_x[1]} \\
    0 - \nabla_{\dot{\mu}_x[0]} \hat F(q;y_i)|_{\dot{\mu}_x[0]} \\
    0 - \nabla_{\dot{\mu}_x[1]} \hat F(q;y_i)|_{\dot{\mu}_x[1]} 
    \end{bmatrix}.
    \label{eq:update rule_2}
\end{equation}

Please note that in our one-layer PC network, when updating $\dot{\mu}_x$, the GM has no expectations regarding acceleration $\ddot{\mu}_x$ and that is why $D$ is designed such that the two 0's appear after the shift. In other words, the GM has no mechanism to estimate the acceleration of the external world, that is, $\frac{\dd}{\dd t}\dot{\mu}_x = \ddot{\mu}_x = (\ddot{\mu}_x[0], \ddot{\mu}_x[1])^\top = (0,0)^\top$. Last but not least, in \textit{line 10} of the pseudo-code, we have used the explicit Runge-Kutta method of order 5(4) (i.e., RK45), implemented in Scipy\footnote{\href{https://docs.scipy.org/doc/scipy/reference/generated/scipy.integrate.RK45.html}{https://docs.scipy.org/doc/scipy/reference/generated/scipy.integrate.RK45.html}} for integrating $\int \frac{\dd}{\dd s}\tilde{\mu}_x^{(s)} \, \dd s$ and updating $\tilde{\mu}_x$.

\begin{algorithm}[ht]
    \caption{Pseudocode for perception modelling as hidden state inference}
    \begin{algorithmic}[1]
    \REQUIRE Observations \(\mathcal{Y} = \{y_0, y_1, \dots, y_{n-1}\}\)
    \ENSURE Inferred \(\tilde\mu_x\) after each observation \(y_i\)
    \STATE Initialise randomly \(\tilde\mu_x=(\mu_x, \dot{\mu}_x)^\top \sim \mathsf{N}(0, I)\)
    \STATE Initialise precision terms: \(\Pi_y = I_{2\times2}\), \(\Pi_x = I_{2\times2}\), $\widetilde\Pi_x = \begin{bsmallmatrix}
     1 &0\\
     0 & 1
\end{bsmallmatrix} \otimes \Pi_x$
    
    \STATE Define state/likelihood flow terms \(f(x, \theta)\) , \(g(x, \theta)\)
        \FOR{\(y_i\) in \(\mathcal{Y}\)}
            \STATE \( \varepsilon_y(\tilde\mu_x) = (y_i - g(\mu_x, \theta))\)
            \STATE \( \varepsilon_x(\tilde\mu_x) =(\dot{\mu}_x - f({\mu}_x, \theta),-\nabla f(\mu_x, \theta)\dot{\mu}_x)^\top\)
            \STATE \(\hat F(q;y_i)=\frac{1}{2}\left[\varepsilon_y(\tilde\mu_x)^\top \Pi_y \varepsilon_y(\tilde\mu_x) + \varepsilon_x(\tilde\mu_x)^\top \tilde\Pi_x \varepsilon_x(\tilde\mu_x)\right]\)
            
            \STATE \(\nabla_{\tilde{\mu}_x} \hat F(q;y_i)=(\nabla_{\mu_x} \hat F(q;y_i), \nabla_{\dot{\mu}_x} \hat F(q;y_i))^\top\)
            \STATE \(\frac{\dd}{\dd t}\tilde{\mu}_x^{(t)} =D\tilde{\mu}_x^{(t)} -\left.\nabla_{\tilde{\mu}_x} \hat F(q;y_i)\right|_{\tilde{\mu}_x^{(t)}}\)

            \STATE \(\tilde{\mu}_x \leftarrow \int \frac{\dd}{\dd s}\tilde{\mu}_x^{(s)} \, \dd s\)

        \ENDFOR
    \end{algorithmic}
    \label{alg:pseudocode}
\end{algorithm}

\section{The Lotka-Volterra GP and observations}\label{app:Lotka-Volterra GP and observations}
\begin{figure}[ht]
    \centering
    \begin{minipage}{0.50\textwidth}
        \centering
        \includegraphics[width=\textwidth]{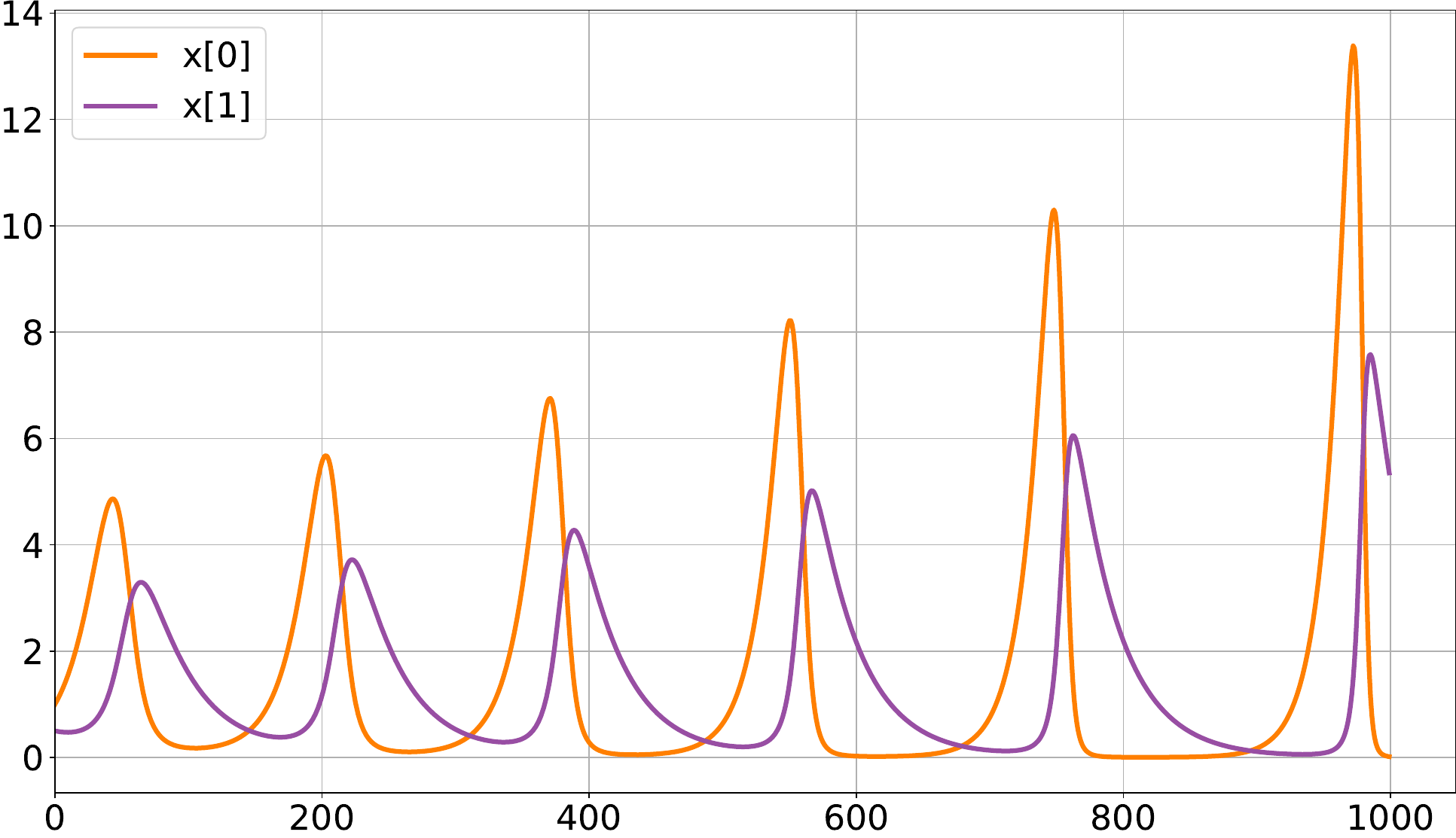}
    \end{minipage}\hfill
    \begin{minipage}{0.495\textwidth}
        \centering
        \includegraphics[width=\textwidth]{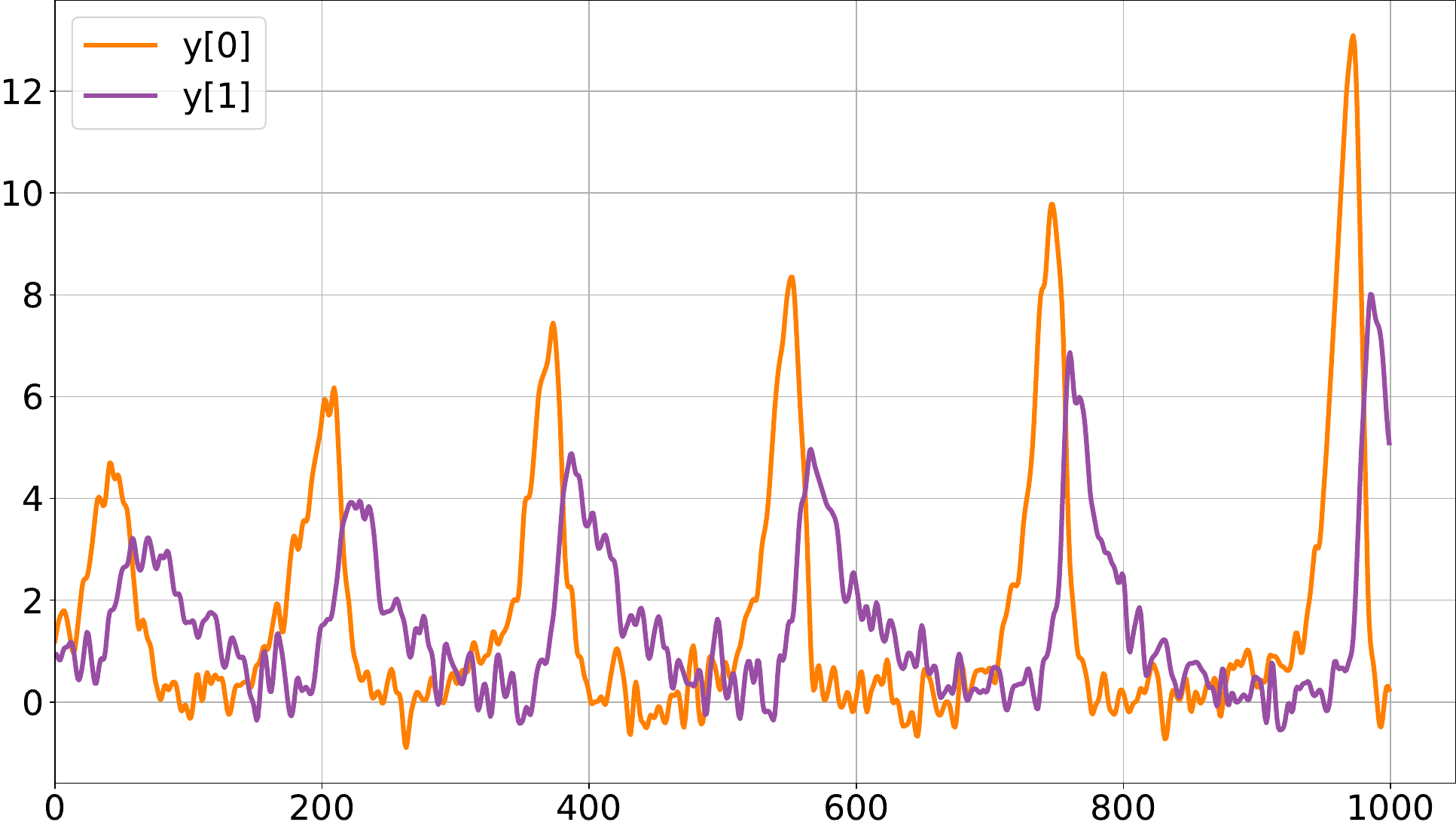}
    \end{minipage}
    \caption{Solution $x$, to the ODEs in the Lotka-Volterra GP (left) and the generated noisy observations $y$ (right).}
    \label{fig:x and y}
\end{figure}

Fig.~\ref{fig:x and y} shows the solution to the Lotka-Volterra GP serving as the hidden state $x$, to be estimated (left) and the generated observations $y$, by adding coloured noise to $x$ (right). 

\section{Variational free energy and variational Laplace}\label{app:VFE functional and mfa}

The true posterior filtering distribution $p(\tilde{x}^t|y^t)$ is approximated by a surrogate distribution $q$ which minimises the variational free energy \eqref{VFE}; recall that $\tilde{x}^t = (x,\dot{x})^\top$. To alleviate notational burden, we omit the $t$ superscript in what follows unless directly relevant. The inference scheme first requires us to put a constraint on the family of distributions $q$ can belong to. To adhere to the predictive coding ethos, the approximating distribution is set to be of the Gaussian family; specifically $q(\tilde x)=p_N(\tilde x; \tilde\mu_x,\Sigma)$, where $\tilde\mu_x = \arg\max_{\tilde{\mu}_x} \ln p(\tilde{\mu}_x, y)$ is the expected posterior mode. To further capitalise on this approximation we employ the \emph{Laplace approximation} on the generative model's posterior density \citep[e.g.][]{FrMa2007}.
The Laplace approximation posits the log-density is approximately quadratic in $\tilde x$ about the mode, $\ln  p(\tilde x, y)\approx \ln p(\tilde\mu_x, y)-\frac{1}{2}\left(\tilde x-\tilde\mu_x\right)^\top U_{\tilde x\tilde x}\left(\tilde x-\tilde\mu_x\right)+\mathrm{const.}$, where $U_{\tilde x\tilde x} = -\frac{\partial^2 \tilde x}{\partial \tilde x^2}\ln  p(\tilde x, y)|_{\tilde x=\tilde \mu_x}$ is the negative curvature. This can be directly applied to simplify the VFE functional to 
\begin{align*}
    F(q;y^t) &= \E_q[-\ln\p(\tilde x, \bfy) + \ln q(\tilde x;\tilde\mu_x,\Sigma)]\\
    &\approx \E_q\left[-\ln p(\tilde\mu_x, y)+\frac{1}{2}\left(\tilde x-\tilde\mu_x\right)^\top U_{\tilde x\tilde x}\left(\tilde x-\tilde\mu_x\right) + \ln q(\tilde x;\tilde\mu_x,\Sigma)\right]\\
    & =-\ln p(\tilde\mu_x, y)+\frac{1}{2}\E_q\left[\left(\tilde x-\tilde\mu_x\right)^\top U_{\tilde x\tilde x}\left(\tilde x-\tilde\mu_x\right)\right] + \E_q\left[\ln q(\tilde x;\tilde\mu_x,\Sigma)\right].
\end{align*}
The first expectation term can be computed by applying matrix identities
\begin{align*}
    \E_q\left[\left(\tilde x-\tilde\mu_x\right)^\top U_{\tilde x\tilde x}\left(\tilde x-\tilde\mu_x\right)\right] &= \mathsf{Tr}\left(\E_q\left[\left(\tilde x-\tilde\mu_x\right)^\top U_{\tilde x\tilde x}\left(\tilde x-\tilde\mu_x\right)\right]\right)\\
    &=\mathsf{Tr}\left(\E_q\left[\left(\tilde x-\tilde\mu_x\right)\left(\tilde x-\tilde\mu_x\right)^\top \right]U_{\tilde x\tilde x}\right)\\
    &=\mathsf{Tr}\left(\Sigma^{-1}U_{\tilde x\tilde x}\right).
\end{align*}
The second expectation term is the differential entropy of a Gaussian random vector and is equal to $\frac{1}{2}\left(d_x \ln 2\pi\e -\ln |\Sigma|\right)$ which, crucially, is free of $\tilde\mu_x$. The variational free energy can be approximated (up to a constant) by 
\[
F(q;y^t) \approx \hat F(q;y^t) = -\ln p(\tilde\mu_x, y) + \frac{1}{2}\mathsf{Tr}\left(\Sigma^{-1}U_{\tilde x\tilde x}\right) -\frac{1}{2}\ln |\Sigma|.
\]
Conditionally on $\tilde\mu_x$, the LHS expression is minimised when $\nabla_\Sigma \hat F(q;y^t) = \frac{1}{2}(U_{\tilde x\tilde x}-\Sigma^{-1})= 0$, which gives the approximate posterior covariance of $\Sigma = U_{\tilde x\tilde x}^{-1}$. 

Thus the VFE  can be approximated (up to a constant) by
\begin{equation}
    \hat F(q;y^t) = \frac{1}{2}\left[\varepsilon_y(\tilde\mu_x)^\top \Pi_y \varepsilon_y(\tilde\mu_x) + \varepsilon_x(\tilde\mu_x)^\top \widetilde\Pi_x \varepsilon_x(\tilde\mu_x)\right],\label{eqn:approx_vfe}
\end{equation}
where $\widetilde\Pi_x = \begin{bsmallmatrix}
     1 &0\\
     0 & 1
\end{bsmallmatrix} \otimes \Pi_x$ and $\otimes$ is the Kronecker product. This expanded precision form arises from the regularisation prior distribution $\nabla f(x)^\top\dot{x}\sim\mathsf{N}(0,\Pi_x)$; this is based on the assumed smoothness of the hidden process, and if higher-order derivatives of $x^t$ (e.g.\ acceleration) were used, a different expanded precision would be required \citep[see e.g.][]{CoMi1965}. 

The approximated VFE \eqref{eqn:approx_vfe} is a functional of $q$  with a value fully expressed through the mode $\tilde\mu_x$. An approximate solution to the inference problem is found by minimising \eqref{eqn:approx_vfe} via gradient-based updates on $\tilde\mu_x$; for details on how this is done see Appendix \ref{app:PC message passing}. The general gradients are given in Appendix \ref{app:gradients} but could also be calculated using auto-differentiation.

\section{Experiment further analysis}\label{app:experiments}
\begin{figure}[ht]
    \centering
    \begin{minipage}{0.49\textwidth}
        \centering
        \includegraphics[width=\textwidth]{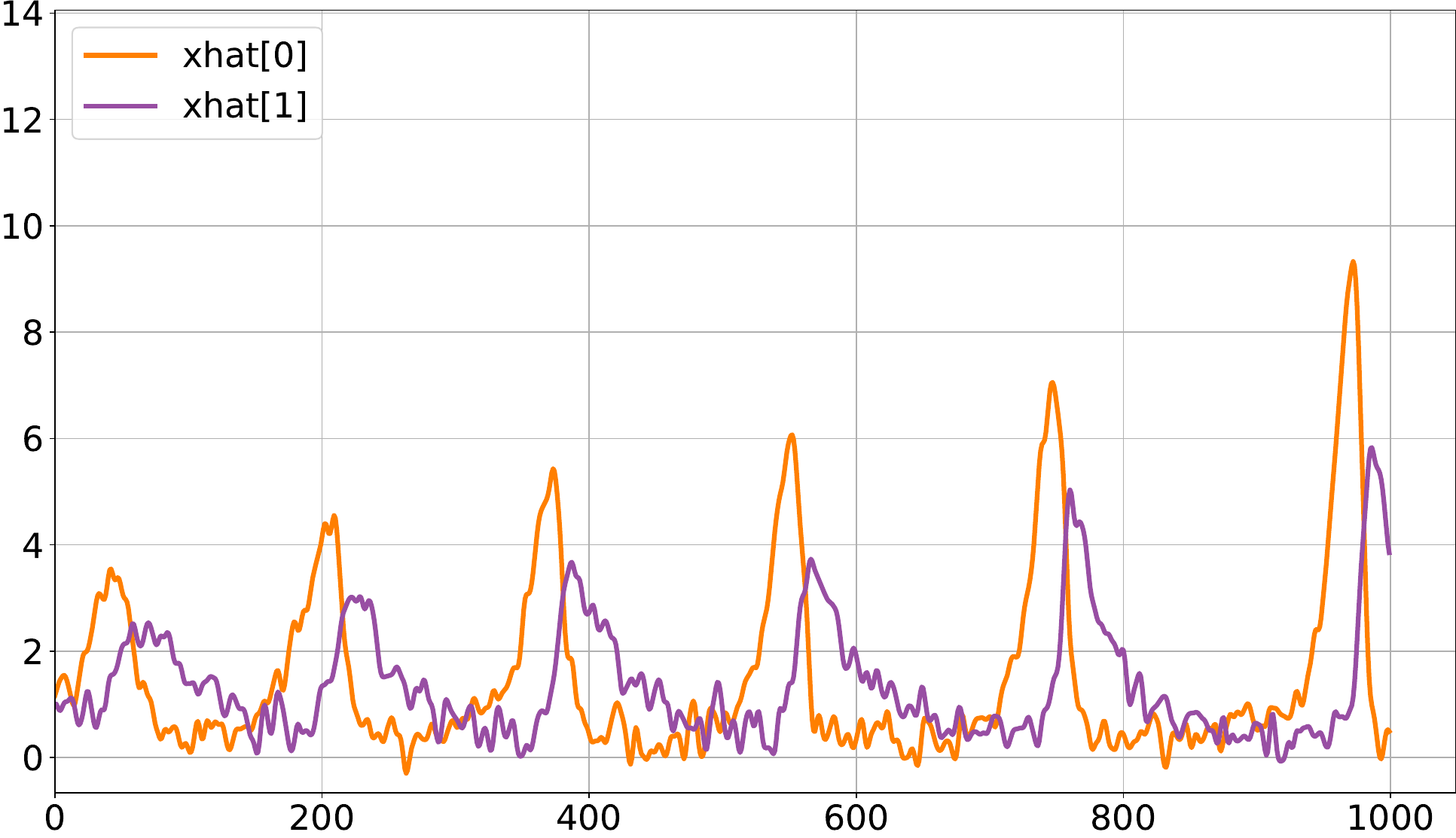}
    \end{minipage}\hfill
    \begin{minipage}{0.49\textwidth}
    \centering
        \includegraphics[width=\textwidth]{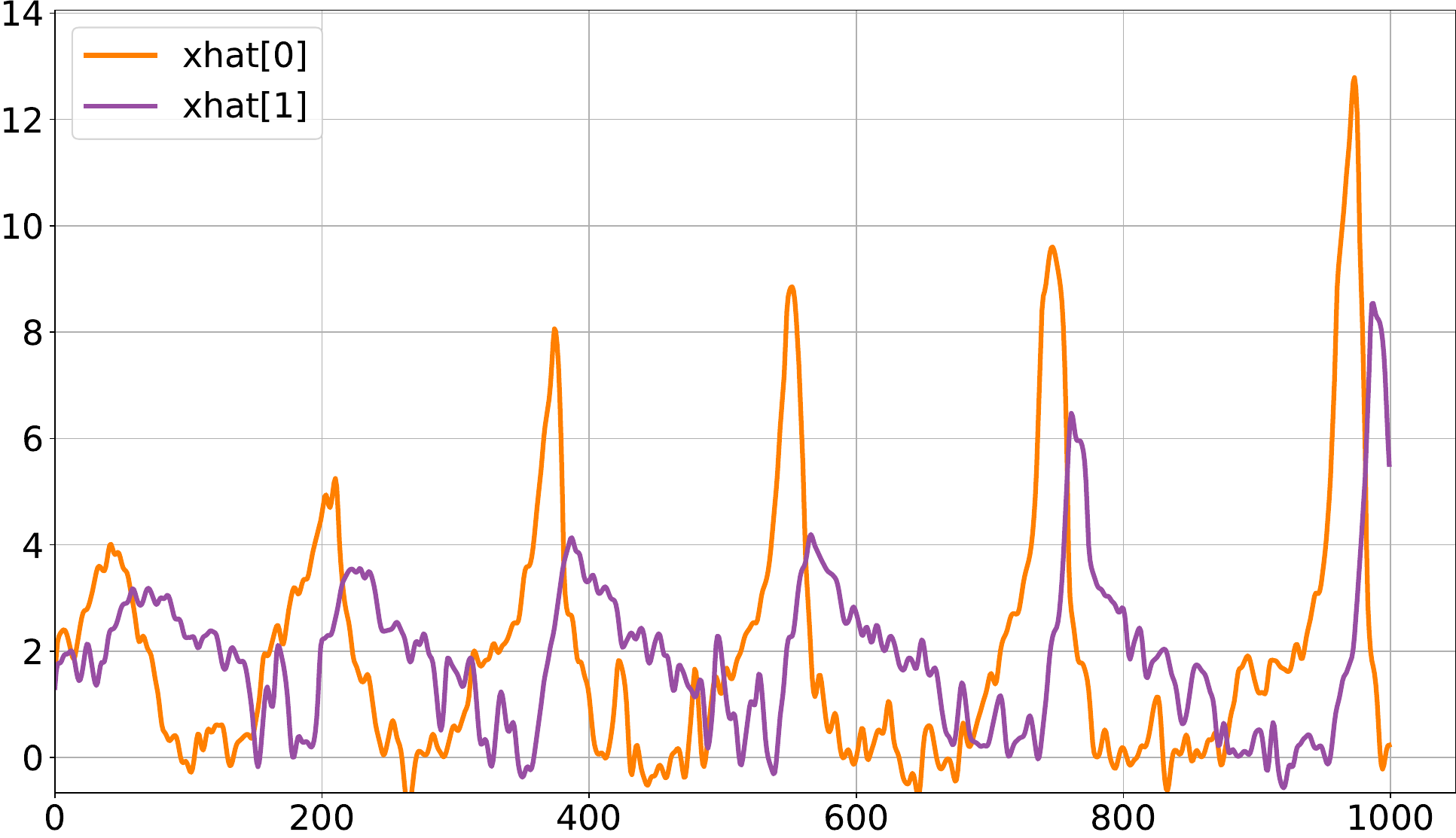}
        
    \end{minipage}
    
    \vspace{0.2cm} 
    
    \begin{minipage}{0.49\textwidth}
        \centering
        \includegraphics[width=\textwidth]{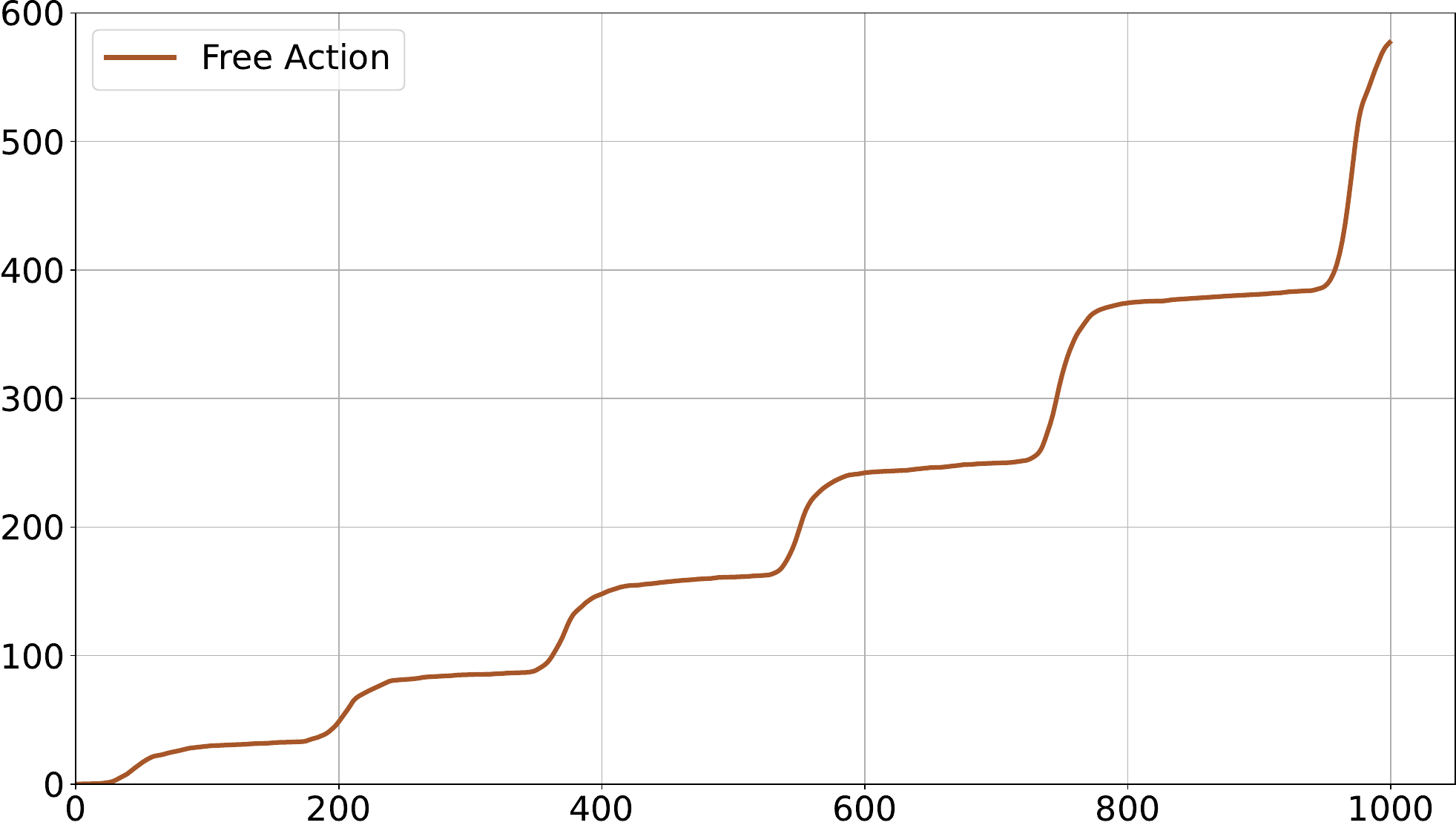}
    \end{minipage}\hfill
    \begin{minipage}{0.49\textwidth}
        \centering
        \includegraphics[width=\textwidth]{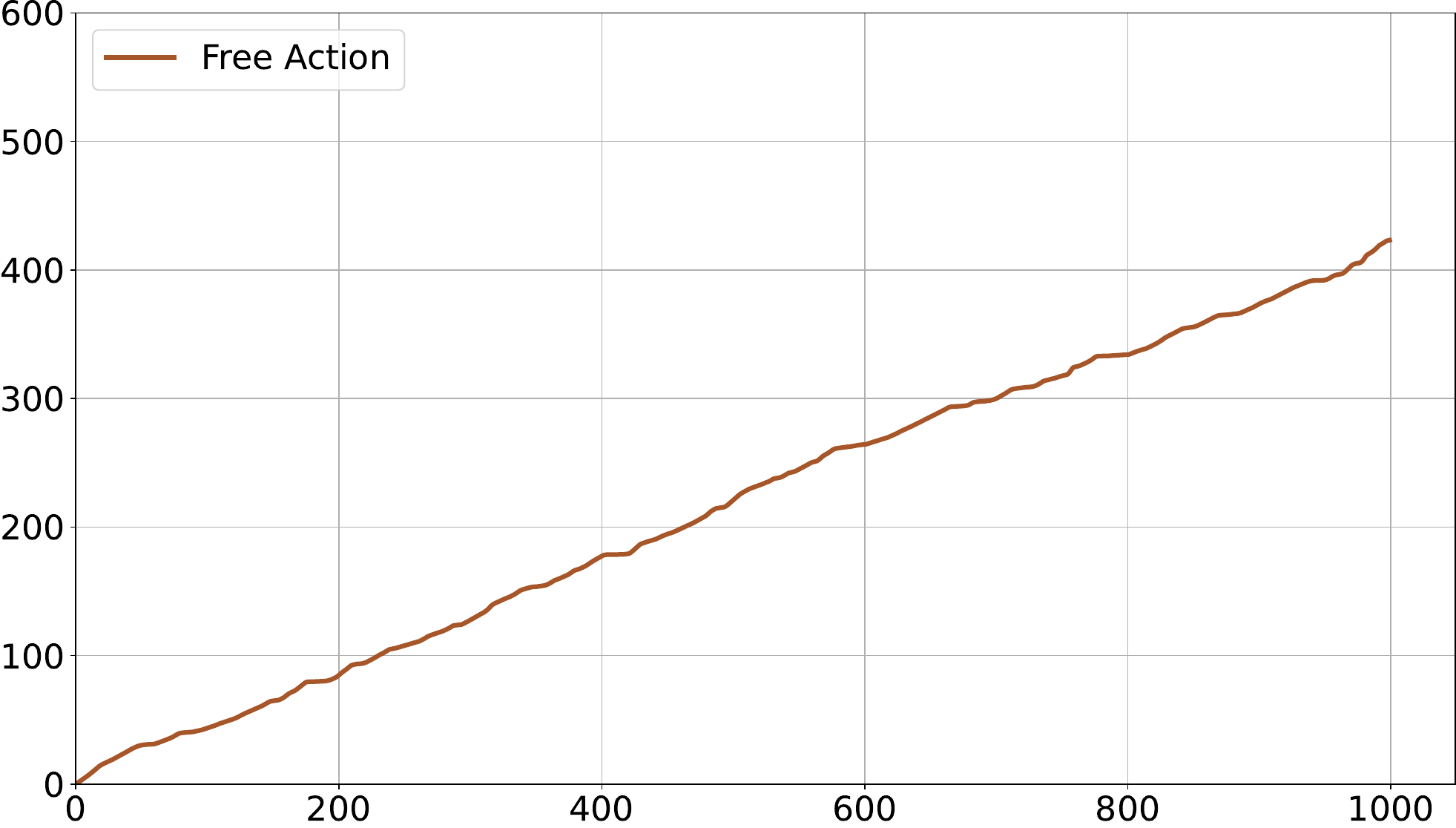}
    \end{minipage}
    
    \caption{Estimated hidden state and free action for $M_1$ in top panel (left) and bottom panel (left), respectively, and estimated hidden state and free action for $M_2$ in top panel (right) and bottom panel (right), respectively.}
    \label{fig:xhat and fa}
\end{figure}

The inferred hidden states and the evolution of free action throughout the inference period is presented in the top and bottom panel of Fig.~\ref{fig:xhat and fa}, for $M_1$ and $M_2$, respectively.

For $M_1$, we have the inferred hidden states, $\hat{x}[0]$ and $\hat{x}[1]$ at top panel (left) along with the evolution of free action during the entire inference period at the bottom panel (left) with a final free action value of 576.98---with MSE error of 0.53. It is very interesting to see that the linear constraint on $f_1(x)$ for $M_1$ forces the free action to have regular steep jumps as if the pullback attractor fails---due to its simple linear form---to capture certain non-linear dynamics of the hidden state at regular intervals given the noisy observations $y[0]$ and $y[1]$.

For $M_2$, we have the inferred hidden states, $\hat{x}[0]$ and $\hat{x}[1]$ at the top panel (right) along with the evolution of free action during the entire inference period at the bottom panel (right) with a final free action value of 423.21---with MSE error of 0.52. It is clear that the trigonometric $f_2(x)$ of model $M_2$, always grows linearly, since it does have the capacity to capture the non-linear dynamics of the hidden states of a Lotka-Volterra generative process much better than the pullback attractor in model $M_1$. We can also see that for $M_2$, the inferred $\hat{x}[0]$ and $\hat{x}[1]$ are much closer to the true $x[0]$ and $x[1]$, compared to $M_1$.

\textbf{Bayesian model comparison/selection}: We can use Bayesian model comparison/selection to decide which model is best. \emph{Bayes Factor} ($\mathsf{BF}$) may be calculated as the ratio of the model evidences---in our case the \emph{free action} is the proxy for the otherwise intractable model evidence---for the two models: $\mathsf{BF}_{1,2}=\frac{P(\mathcal{Y}|M_1)}{P(\mathcal{Y}|M_2)} \approx \frac{\mathsf{FA}(\mathcal{Y}|M_1)}{\mathsf{FA}(\mathcal{Y}|M_2)}=\frac{576.98}{423.21}=1.36$, where $\mathsf{FA}$ denotes the free action discussed in Section \ref{various problem formulations and their implications}. Since $\mathsf{BF} > 1.0$, we choose $M_2$.

Additionally, we can see that the MSE error of $M_1$ and $M_2$ are almost identical, that is, their fitting power (i.e., inference accuracy) is almost the same. Crucially, it should be noted that even if the MSE error of $M_2$ was larger than that of $M_1$, we should still pick $M_2$ as the better model. This is because choosing a model merely based on its fitting power (i.e.\ highest accuracy/lowest MSE) with no regards for model complexity, puts the GM at the risk of overfitting the data and failing to generalise, which is a common problem in the world of machine learning and AI.

\section{The generative power of a one-layer PC network}\label{app:generative_power}
\begin{figure}[ht]
    \centering
    \begin{minipage}{0.49\textwidth}
        \centering
        \includegraphics[width=\textwidth]{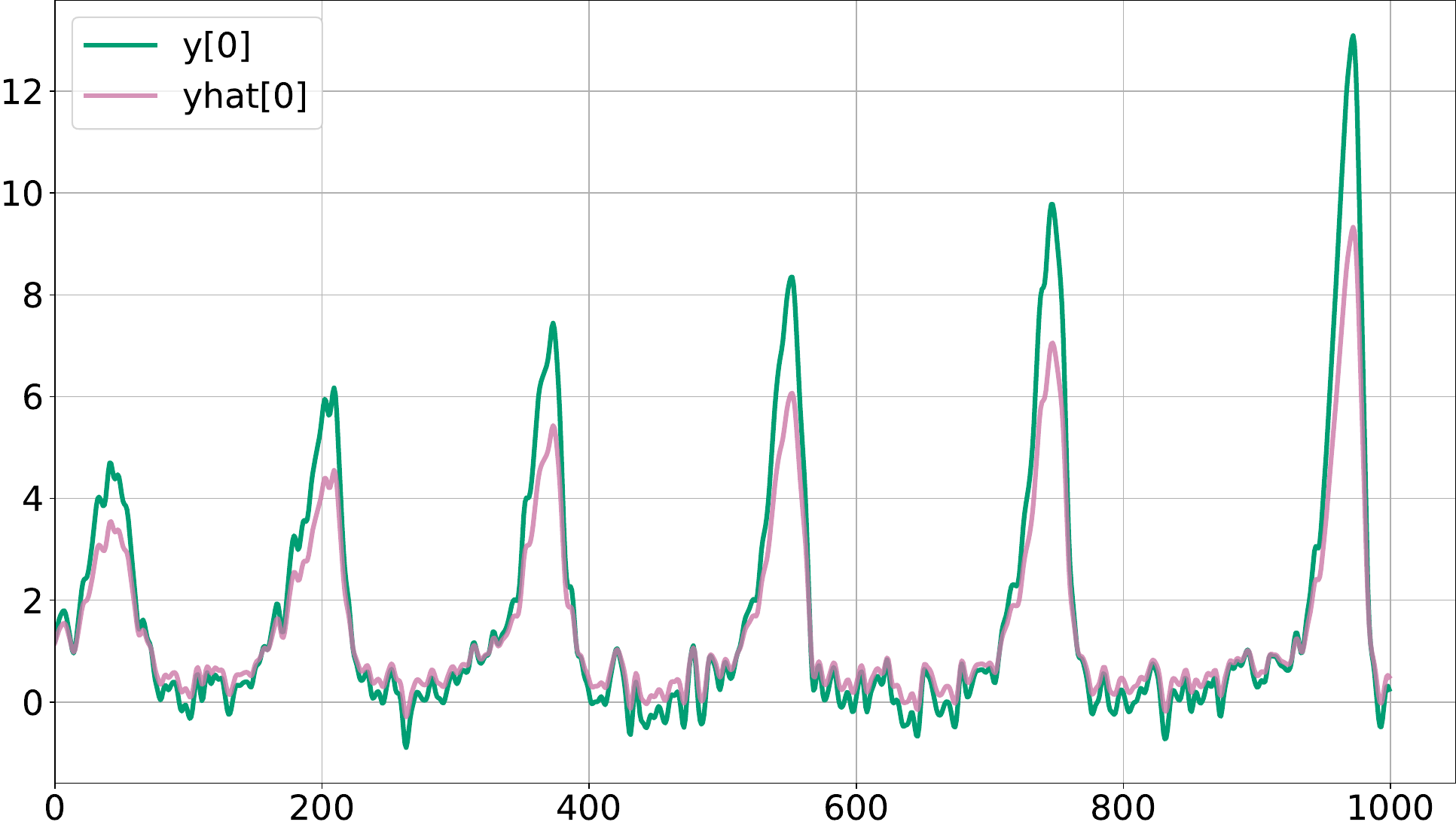}
    \end{minipage}\hfill
    \begin{minipage}{0.49\textwidth}
        \centering
        \includegraphics[width=\textwidth]{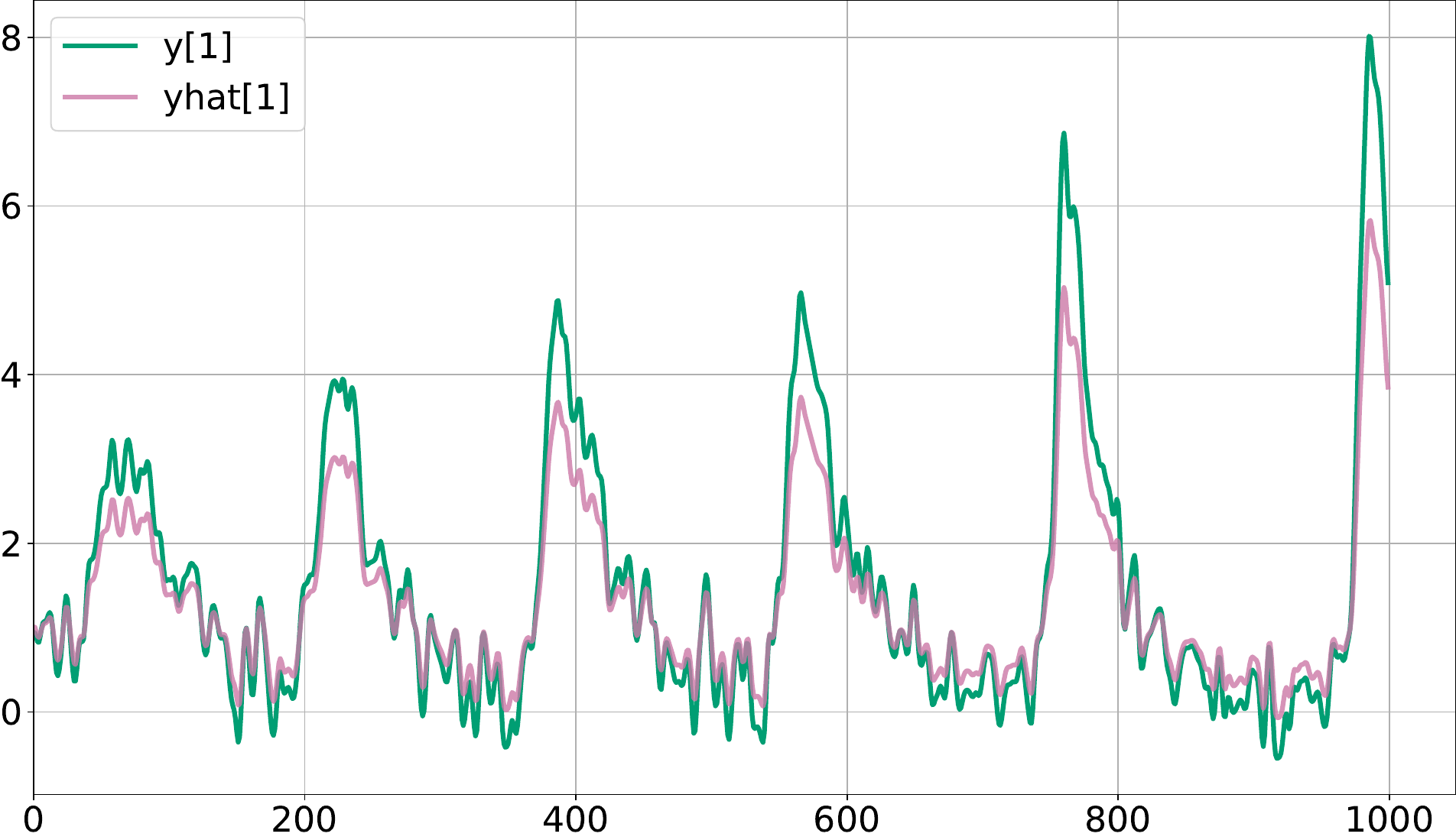}
    \end{minipage}
    
    \vspace{0.2cm} 
    
    \begin{minipage}{0.49\textwidth}
        \centering
        \includegraphics[width=\textwidth]{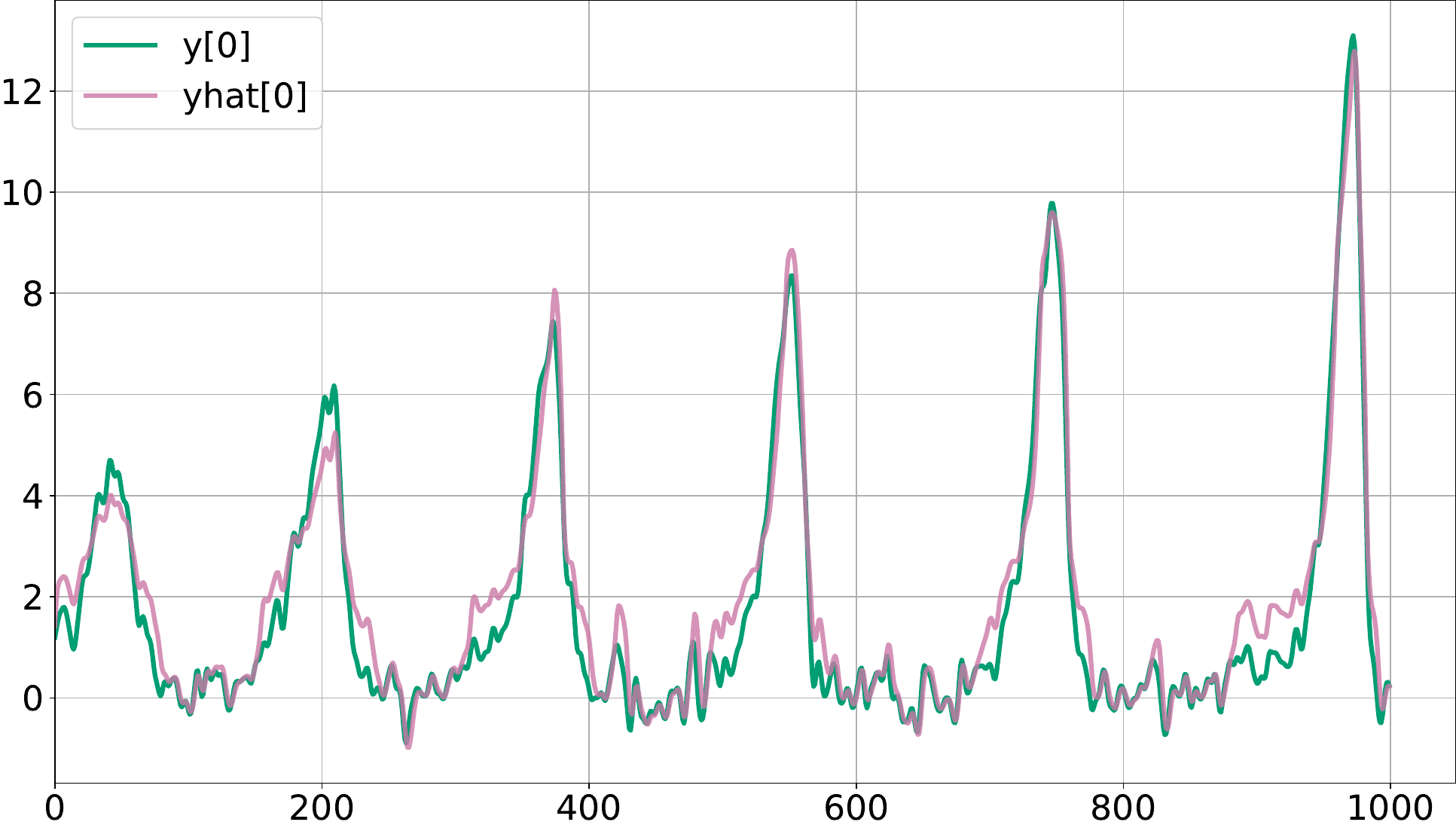}
    \end{minipage}\hfill
    \begin{minipage}{0.49\textwidth}
        \centering
        \includegraphics[width=\textwidth]{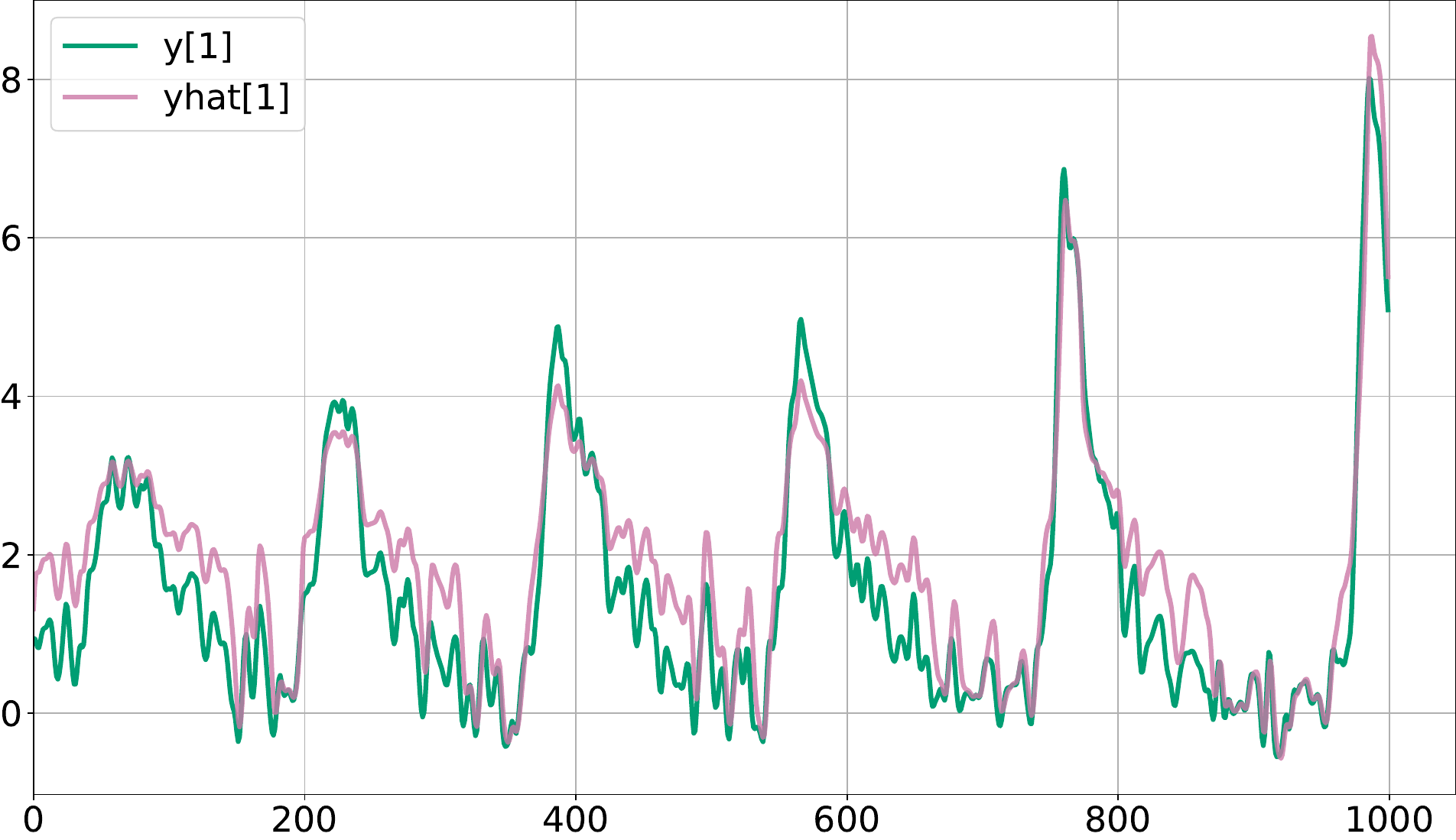}
    \end{minipage}
    
    \caption{Predicted sensations and actual sensations for model $M_1$ (top panel) and model $M_2$ (bottom panel).}
    \label{fig:y and yhat}
\end{figure}

The one-layer PC network is a generative model and as such, it has the ability to generate data using the $g(\mu_x)$ term, that maps the expected state of the world $\mu_x$ to its expected observation $\hat{y}$, at any given time $t$. Note that the true sensations $y$ have 2 dimensions $y[0]$ and $y[1]$ and as such the GM will generate predictions for both dimensions, denoted as $\hat{y}[0]$ and $\hat{y}[1]$. The top panel of Fig.~\ref{fig:y and yhat}, shows the generative power of model $M_1$, where the predicted sensations $\hat{y}[0]$ and $\hat{y}[1]$ are plotted against the \textit{actual} received sensations $y[0]$ and $y[1]$. The bottom panel shows the same but for model $M_2$. We can see that $M_2$ makes more accurate predictions, especially when it comes to matching the high magnitudes of the true sensations.

\newpage
\section{Calculating the gradients of the approximate VFE for a given generative model}\label{app:gradients}
The VFE for the generative models considered in the main paper can be approximated by $\hat F(q;y^t)=\frac{1}{2}\left[\varepsilon_y(\tilde\mu_x)^\top \Pi_y \varepsilon_y(\tilde\mu_x) + \varepsilon_x(\tilde\mu_x)^\top \widetilde\Pi_x \varepsilon_x(\tilde\mu_x)\right]$, where $\varepsilon_x(\tilde\mu_x) =\left(
   \dot{\mu}_x - f({\mu}_x),-\nabla f(\mu_x)\dot{\mu}_x
\right)^\top$
and $\varepsilon_y(\tilde\mu_x) = y - g(\mu_x)$; the Jacobian term $\nabla f(\mu_x)$ arising from the regularising prior is treated as constant for inference purposes \citep[e.g.][]{HeMi2024}. Differentiating the functional w.r.t.\ components of $\tilde\mu_x$, we get 
\begin{align*}
    \nabla_{\mu_x} \hat F(q;y^t) &= [\nabla_{\mu_x}\varepsilon_y(\tilde\mu_x)]^\top \Pi_y \varepsilon_y(\tilde\mu_x) + [\nabla_{\mu_x}\varepsilon_x(\tilde\mu_x)]^\top \widetilde\Pi_x \varepsilon_x(\tilde\mu_x)\\
    &= -[\nabla g(\mu_x)]^\top \Pi_y [y - g(\mu_x)] -[\nabla f(\mu_x)]^\top \Pi_x [\dot{\mu}_x - f({\mu}_x)],
\end{align*}
and
\begin{align*}
    \nabla_{\dot{\mu}_x} \hat F(q;y^t) &= [\nabla_{\dot{\mu}_x}\varepsilon_y(\tilde\mu_x)]^\top \Pi_y \varepsilon_y(\tilde\mu_x) + [\nabla_{\dot{\mu}_x}\varepsilon_x(\tilde\mu_x)]^\top \widetilde\Pi_x \varepsilon_x(\tilde\mu_x)\\
    &=\Pi_x [\dot{\mu}_x - f({\mu}_x)]+[-\nabla f(\mu_x)]^\top \Pi_x [-\nabla f(\mu_x)\dot{\mu}_x].
\end{align*}
The Jacobian terms $\nabla f(\mu_x)$ and $\nabla g(\mu_x)$ are obtained by differentiating the vector-valued functions w.r.t.\ $x$ and evaluating at the estimates $\mu_x$; that is, $\nabla f(\mu_x) = \left.\nabla f(x)\right|_{x=\mu_x}$ and $\nabla g(\mu_x) = \left.\nabla g(x)\right|_{x=\mu_x}$. 

The above gradients can then be used in an optimisation scheme for minimising the variational free energy; for example, through a gradient descent algorithm with momentum, as outlined in Appendix \ref{app:PC message passing}.

\end{document}

%% file: macros.tex
\newcommand{\E}{\mathbb{E}}

\newcommand{\dd}{\mathsf{d}}

\newcommand{\bfy}{{\mathbf{y}}}

\newcommand{\p}{\mathsf{p}}

\newcommand{\Real}{\mathbb{R}}

\newcommand{\e}{\mathsf{e}}

\renewcommand{\max}{\mathsf{max}}

\renewcommand{\arg}{\mathsf{arg}}

